\documentclass[sigconf]{acmart}

\usepackage{amsmath}
\usepackage{booktabs}
\usepackage{graphicx}
\usepackage{array}

\graphicspath{{Materials/}}

\copyrightyear{2026}
\acmYear{2026}
\setcopyright{cc}
\setcctype{by-nc-nd}
\acmConference[SIGSPATIAL '26]{The 34th ACM International Conference on Advances in Geographic Information Systems}{November 03--06, 2026}{Riverside, CA, USA}
\acmBooktitle{The 34th ACM International Conference on Advances in Geographic Information Systems (SIGSPATIAL '26), November 03--06, 2026, Riverside, CA, USA}
\acmDOI{10.1145/3841645.3843383}
\acmISBN{979-8-4007-2950-8/2026/11}

\begin{document}

\title[BEACON]{BEACON: Behavioral and Semantic Enrichment of AlphaEarth Embeddings through Tri-Modal Contrastive Learning}

\author{Hao Tian}
\affiliation{%
  \department{Department of Geography}
  \institution{Texas A\&M University}
  \city{College Station}
  \state{TX}
  \country{USA}}
\email{haotian@tamu.edu}          

\author{Heng Cai}
\authornote{Corresponding author.}
\affiliation{%
  \department{Department of Geography}
  \institution{Texas A\&M University}
  \city{College Station}
  \state{TX}
  \country{USA}}
\email{hengcai@tamu.edu}             

\author{Yifan Yang}
\affiliation{%
  \department{Department of Geography}
  \institution{Texas A\&M University}
  \city{College Station}
  \state{TX}
  \country{USA}}
\email{yyang295@tamu.edu} 

\renewcommand{\shortauthors}{Tian et al.}

\begin{abstract}
Geospatial foundation models such as the AlphaEarth Foundation produce compact and globally consistent representations of the Earth's surface that transfer effectively to a wide range of downstream tasks. However, because these models are trained primarily on Earth-observation imagery, their embeddings mainly capture physical and spectral characteristics while encoding human activity and urban function only weakly. To address this limitation, we propose BEACON, a tri-modal contrastive learning framework that aligns three complementary views of urban space: physical representations from AE embeddings, semantic representations from point-of-interest (POI) text, and human behavioral representations from hourly POI visitation, while keeping the deployed representation image-only. Using the Houston Metropolitan Area as a case study area, we evaluated the performance of the BEACON framework on nine downstream tasks, including seven regression and two classification tasks against six baselines (raw coordinates, Space2Vec, SatCLIP, TESSERA, Clay and AlphaEarth), using frozen linear and MLP probes over five seeds. Under a linear probe, BEACON improves relative $R^2$ over AlphaEarth by up to 43\% for obesity prevalence, 34\% for poor mental health, and 22\% for median household income, while remaining competitive in the prediction of physical and environmental variables. These findings highlight the value of augmenting geospatial foundation models with semantic and behavioral signals, extending their applicability from physical Earth observation to human-centered urban analytics.
\end{abstract}


\maketitle

\section{Introduction}
Geospatial foundation models are changing the way in which urban and environmental prediction tasks are approached. Rather than constructing task-specific covariates for every outcome, analysts can now start from transferable spatial embeddings learned from large-scale Earth-observation data. This is especially useful in urban studies, where labels are sparse, spatial supports vary across agencies, and the same representation may need to be aggregated from pixels to parcels, census units, and regular grids. Recent geospatial foundation-style models, including SatCLIP, TESSERA, Clay, Prithvi and AlphaEarth, show strong transfer across remote-sensing and mapping tasks. AlphaEarth is particularly relevant because it provides a globally consistent annual embedding field at 10 m resolution \cite{brown2025alphaearth}.

However, many urban phenomena are governed by functional and behavioral processes that are not directly observable from overhead imagery. Socioeconomic conditions, public health outcomes, and human activity patterns emerge from interactions among the built environment, institutional functions, and mobility behaviors. This raises a critical question for geospatial foundation models: do image-derived embeddings capture the functional and behavioral organization of urban space, or primarily its physical morphology?

Physical appearance and urban function are often partially aligned. Visually similar locations may perform very different social and economic roles, such as a medical office, a retail center, a school, or a government facility embedded within comparable built forms. Meanwhile, visually distinct locations may exhibit similar behavioral or functional profiles when serve comparable roles within the urban system. Consequently, models trained solely on Earth-observation imagery may have insufficient capacity to distinguish places according to their social, economic, and behavioral characteristics. Such limitations constrain their effectiveness for human-centered urban analytics, where accurate representation of urban function and behavior is essential for understanding accessibility, equity, public health, and socioeconomic dynamics.

Point-of-Interests (POIs) information provide human-centered location semantics by recording the functional, social and contextual meaning associated with a geographic location. Prior work uses POI embeddings to infer urban functional use \cite{niu2021urbanfunctional}, and recent language-location pretraining aligns urban space with POI descriptions \cite{wang2025poi}. AETHER is the closest predecessor to our work: it adapts AlphaEarth through POI-guided contrastive learning and shows that Earth-observation embeddings can be enriched with human-centered urban semantics \cite{liu2025beyondalphaearth}. However, POI text mainly describes intended or registered function. It does not directly tell us how strongly a place is used, when people visit it, or whether its temporal role differs from visually similar places.

Human mobility provides a complementary behavioral view. Hourly and seasonal visitation profiles reveal commuter peaks, weekend leisure, late-night activity, routine service visits, and seasonal demand. MoRA argues that mobility can serve as a backbone for large-scale geospatial representation learning \cite{wen2026mora}. Our goal is more targeted: rather than replacing AlphaEarth with a mobility backbone, we use POI-level mobility as supervision to human-contextualize an already strong physical embedding field.

We introduce BEACON, a tri-modal contrastive learning framework that human-contextualizes AlphaEarth embeddings by integrating location semantics and human behavioral signals into earth-observation representations. BEACON uses POIs as anchors and aligns three views of urban space: AlphaEarth embeddings representing physical context, textual POI descriptions, and temporal visitation profiles. BEACON aligns AlphaEarth embeddings with POI semantics and mobility behavior during training. Once trained, however, the model produces enriched representations from AlphaEarth inputs alone, without requiring POI or mobility data at inference time. Our contributions are: (1) a lightweight tri-modal contrastive objective for adapting AlphaEarth with semantic and behavioral supervision; (2) a Houston metropolitan benchmark spanning socioeconomic, health, activity, environmental, land-cover, and land-use prediction tasks; and (3) evidence that human-contextualized embeddings improve human-centered prediction while largely preserving physical mapping performance.

\section{Data and Methodology}

\begin{figure}[!htbp]
  \centering
  \includegraphics[width=\columnwidth]{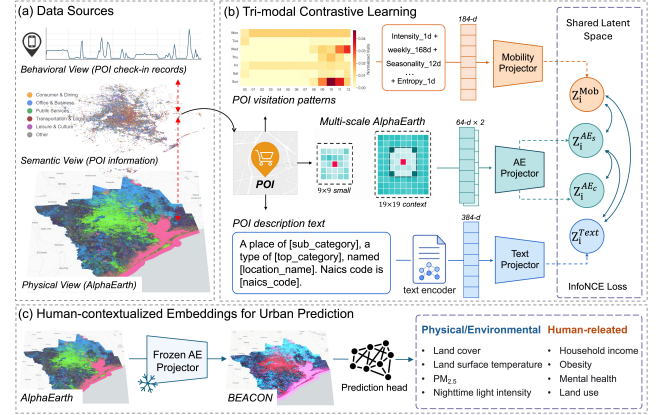}
  \Description{Schematic of the BEACON framework: AlphaEarth physical embeddings,
  POI semantic attributes, and hourly mobility profiles are encoded into a shared
  space and aligned by tri-modal contrastive learning.}
  \caption{Overview of BEACON. Physical, semantic, and behavioral views are aligned through tri-modal contrastive learning to produce human-contextualized embeddings.}
  \label{fig:framework}
\end{figure}

Figure~\ref{fig:framework} illustrates the overall BEACON framework. Panel (a) visualizes the three complementary views of urban space. Panel (b) shows the tri-modal contrastive learning framework, where AlphaEarth embeddings, POI description text, and mobility representations are projected into a shared latent space and aligned using an InfoNCE objective. Panel (c) demonstrates the downstream application workflow. After pretraining, the frozen AlphaEarth projector generates BEACON embeddings from AlphaEarth inputs, which are subsequently evaluated across a diverse set of physical/environmental and human-centered urban prediction tasks.

\subsection{Study Area and Data}
All experiments are conducted in the Houston–The Woodlands–Sugar Land Metropolitan Statistical Area using data from 2024. AlphaEarth embeddings serve as the physical representation backbone, while POI attributes and hourly visitation patterns provide semantic and behavioral supervision during pretraining. Downstream evaluation spans socioeconomic, health, environmental, land-cover, and land-use tasks (Table~\ref{tab:tasks}), enabling assessment of both human-centered and physical representations of urban space.

\begin{table}[!htbp]
  \centering
  \small
  \caption{Downstream benchmark tasks, spatial supports, and common-support sample sizes.}
  \label{tab:tasks}
  \begin{tabular}{@{}llrl@{}}
    \toprule
    Target & Support & $N$ & Task \\
    \midrule
    Median household income & Block groups & 4{,}174 & Reg. \\
    Obesity prevalence & Tracts & 1{,}613 & Reg. \\
    Poor mental health & Tracts & 1{,}613 & Reg. \\
    Nighttime lights & 500\,m px & 139{,}186 & Reg. \\
    LST daytime & 1\,km px & 27{,}424 & Reg. \\
    LST nighttime & 1\,km px & 27{,}424 & Reg. \\
    PM$_{2.5}$ & 0.01$^\circ$ px & 24{,}338 & Reg. \\
    Land cover (15 classes)& 10\,m samples & 298{,}944 & Cls. \\
    Functional land use (10 classes)& Parcels & 255{,}431 & Cls. \\
    \bottomrule
  \end{tabular}
\end{table}

\subsection{AlphaEarth Encoder}
AlphaEarth provides a 64-dimensional embedding for each 10\,m grid cell. For each POI $i$, we extract two complementary AlphaEarth spatial views centered on the POI~\cite{liu2025beyondalphaearth}. Let $a_p \in \mathbb{R}^{64}$ be the embedding at grid cell $p$. We define a smaller view over a local neighborhood $\Omega_i^{(s)}$ (a $9\times9$ window) and a larger context view over a larger neighborhood $\Omega_i^{(c)} \supset \Omega_i^{(s)}$ (a $19\times19$ window), via average pooling:
\begin{equation}
a_i^{(s)} = \tfrac{1}{|\Omega_i^{(s)}|}\!\!\sum_{p \in \Omega_i^{(s)}}\!\! a_p,\quad
a_i^{(c)} = \tfrac{1}{|\Omega_i^{(c)}|}\!\!\sum_{p \in \Omega_i^{(c)}}\!\! a_p .
\end{equation}
Both pooled embeddings are projected into the shared latent space using a shared projector and subsequently $\ell_2$-normalized. The resulting embeddings are denoted by
$z_i^{AE_s}$ and $z_i^{AE_c}$.

\subsection{POI Description Encoder}
Let $s_i,c_i,n_i,k_i$ denote the sub-category, top-level category, name, and NAICS code of POI $i$. We form a textual description with the template ``\emph{A place of $s_i$, a type of $c_i$, named $n_i$. The NAICS code is $k_i$}'' (omitting missing attributes), and denote the resulting description by $t_i$. A pretrained encoder $E(\cdot)$ (OpenAI \texttt{text-embedding-3-large}) produces a 384-dimensional semantic embedding, which a text projector $g_\phi(\cdot)$ maps into the shared space:
\begin{equation}
z_i^{Text} = \mathrm{Norm}\!\left(g_\phi(E(t_i))\right) .
\end{equation}
The location semantic view supplies functional context that is often not observable from overhead imagery, helping to distinguish visually similar locations with different urban functions.

\subsection{Mobility Pattern Encoder}

For each POI $i$, we construct a 184-dimensional mobility profile $m_i$ by concatenating a normalized hour-of-week pattern $w_i\in\mathbb{R}^{168}$, a normalized monthly pattern $o_i\in\mathbb{R}^{12}$, and four scalar descriptors $\psi_i\in\mathbb{R}^{4}$:

\begin{equation}
m_i=[w_i;o_i;\psi_i]\in\mathbb{R}^{184}.
\end{equation}

The descriptor vector $\psi_i$ contains four summary statistics: log total visits, Shannon entropy of the hour-of-week distribution, weekday--weekend gap, and monthly coefficient of variation. The entropy is computed as

\begin{equation}
H(w_i)=-\sum_k w_{ik}\log w_{ik}.
\end{equation}

We retain only POIs with complete mobility records throughout 2024, resulting in 160,944 POIs for representation learning. The mobility profile is mapped into the shared latent space using a two-layer MLP ($184\!\rightarrow\!256\!\rightarrow\!128$, GELU) followed by $\ell_2$ normalization to produce the mobility embedding $z_i^{Mob}$.

\subsection{Tri-Modal Contrastive Alignment}
For a minibatch of $N$ POIs, each sample yields four $\ell_2$-normalized embeddings: $z_i^{AE_s}$, $z_i^{AE_c}$, $z_i^{Text}$, and $z_i^{Mob}$. For any two modalities $u,v$ we use a symmetric CLIP-style objective, where matched POIs form positive pairs:
\begin{equation}
\mathcal{L}(u,v) = \tfrac{1}{2}\!\left[\mathrm{CE}\!\left(\tfrac{Z^{u}(Z^{v})^{\top}}{\tau},y\right) + \mathrm{CE}\!\left(\tfrac{Z^{v}(Z^{u})^{\top}}{\tau},y\right)\right],
\end{equation}
with $Z^u,Z^v\in\mathbb{R}^{N\times d}$, temperature $\tau$, and labels $y=(1,\dots,N)$. The overall objective combines four alignment terms:
\begin{equation}
\begin{aligned}
\mathcal{L}_{\text{total}} =\;& \lambda_{st}\,\mathcal{L}(z^{AE_s},z^{Text}) + \lambda_{sm}\,\mathcal{L}(z^{AE_s},z^{Mob}) \\
& + \lambda_{tm}\,\mathcal{L}(z^{Text},z^{Mob}) + \lambda_{sc}\,\mathcal{L}(z^{AE_s},z^{AE_c}).
\end{aligned}
\end{equation}
We set $\tau=0.07$ and $(\lambda_{st},\lambda_{sm},\lambda_{tm},\lambda_{sc})=(0.35,0.35,0.15,0.15)$, and train with AdamW (learning rate $1\!\times\!10^{-3}$, weight decay $1\!\times\!10^{-4}$, cosine schedule, batch size 512, 100 epochs) on an 80/10/10 POI split.

\subsection{Inference and Downstream Evaluation}
After pretraining, the frozen AE projector is applied to the full AlphaEarth embedding field, transforming each 64-dimensional embedding into a 128-dimensional BEACON representation. Regional embeddings are obtained by mean pooling over target spatial units. We compare BEACON against six baselines (coordinates, Space2Vec, SatCLIP, TESSERA, Clay v1.5, and AlphaEarth) using frozen embeddings with linear and single-hidden-layer MLP probes. Performance is evaluated using five random 70/10/20 train-validation-test splits, reporting mean $\pm$ standard deviation of $R^2$ for regression and Macro-F1 for classification tasks.

\section{Results}
\begin{table*}[!htbp]
  \centering
  \small
  \caption{Regression performance ($R^2$, linear probe) across seven downstream tasks (mean\,$\pm$\,s.d.\ over five seeds). \textbf{Best} in bold, \underline{second-best} underlined.}
  \label{tab:reg}
  \setlength{\tabcolsep}{4.5pt}
  \begin{tabular}{@{}l cccc ccc@{}}
    \toprule
    & \multicolumn{4}{c}{Human-related} & \multicolumn{3}{c}{Physical / Environmental} \\
    \cmidrule(lr){2-5}\cmidrule(lr){6-8}
    Model & Income & Obesity & Mental health & NTL & LST day & LST night & PM$_{2.5}$ \\
    \midrule
    Coordinates & 0.012\,$\pm$\,0.007 & 0.032\,$\pm$\,0.020 & 0.040\,$\pm$\,0.021 & 0.002\,$\pm$\,0.000 & 0.246\,$\pm$\,0.009 & 0.304\,$\pm$\,0.006 & 0.110\,$\pm$\,0.004 \\
    Space2Vec   & 0.280\,$\pm$\,0.027 & 0.472\,$\pm$\,0.038 & 0.328\,$\pm$\,0.060 & 0.419\,$\pm$\,0.071 & 0.464\,$\pm$\,0.014 & 0.331\,$\pm$\,0.009 & 0.386\,$\pm$\,0.007 \\
    SatCLIP     & 0.024\,$\pm$\,0.010 & 0.062\,$\pm$\,0.025 & 0.047\,$\pm$\,0.020 & 0.316\,$\pm$\,0.050 & 0.644\,$\pm$\,0.009 & 0.584\,$\pm$\,0.003 & 0.617\,$\pm$\,0.009 \\
    TESSERA     & \underline{0.420\,$\pm$\,0.027} & 0.519\,$\pm$\,0.046 & \underline{0.431\,$\pm$\,0.041} & \underline{0.667\,$\pm$\,0.111} & 0.824\,$\pm$\,0.009 & 0.779\,$\pm$\,0.008 & \underline{0.868\,$\pm$\,0.004} \\
    Clay        & 0.413\,$\pm$\,0.034 & \underline{0.557\,$\pm$\,0.048} & 0.429\,$\pm$\,0.031 & 0.603\,$\pm$\,0.087 & \textbf{0.860\,$\pm$\,0.004} & \textbf{0.813\,$\pm$\,0.007} & \textbf{0.905\,$\pm$\,0.002} \\
    AlphaEarth     & 0.379\,$\pm$\,0.031 & 0.431\,$\pm$\,0.040 & 0.393\,$\pm$\,0.053 & 0.663\,$\pm$\,0.112 & \underline{0.840\,$\pm$\,0.007} & \underline{0.792\,$\pm$\,0.005} & 0.845\,$\pm$\,0.004 \\
    \textbf{BEACON} & \textbf{0.463\,$\pm$\,0.025} & \textbf{0.614\,$\pm$\,0.035} & \textbf{0.528\,$\pm$\,0.022} & \textbf{0.680\,$\pm$\,0.114} & 0.843\,$\pm$\,0.006 & 0.787\,$\pm$\,0.006 & 0.856\,$\pm$\,0.003 \\
    \bottomrule
  \end{tabular}
\end{table*}

\subsection{Pretraining Dynamics and Cross-Modal Alignment}
\begin{figure}[!htbp]
  \centering
  \includegraphics[width=\columnwidth]{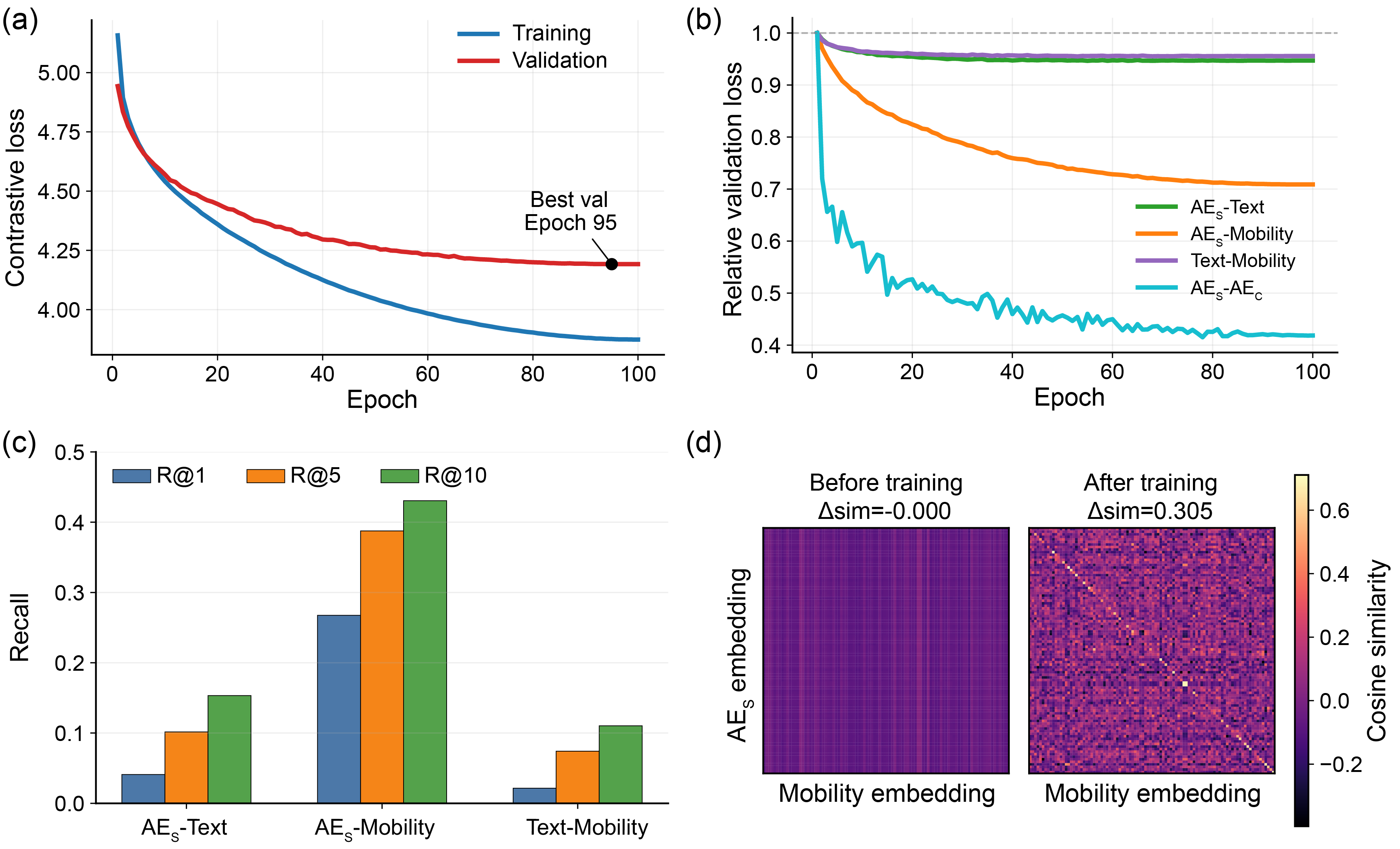}
  \Description{Four-panel figure of BEACON pretraining diagnostics: contrastive
  loss curves, per-term relative validation loss, cross-modal retrieval recall,
  and AlphaEarth--mobility cosine similarity before and after training.}
  \caption{Pretraining dynamics of BEACON. (a) Training/validation contrastive loss. (b) Relative validation loss of the four objective terms. (c) Held-out cross-modal retrieval Recall@1/5/10. (d) AlphaEarth--mobility cosine-similarity before/after training.}
  \label{fig:training}
\end{figure}

The tri-modal objective converges stably with little evidence of overfitting (Figure~\ref{fig:training}a). Analysis of the validation loss components (Figure~\ref{fig:training}b) shows that alignment is driven primarily by the intra-modal AE$_s$--AE$_c$ term and the cross-modal AE$_s$--Mobility term, while text-based alignments contribute more modestly. Consistent with this observation, cross-modal retrieval results (Figure~\ref{fig:training}c) indicate that mobility provides the strongest supervisory signal among the human-centered modalities. The similarity matrices further reveal a substantial increase in AlphaEarth--mobility correspondence after training (Figure~\ref{fig:training}d). Together, these results demonstrate that BEACON effectively injects human-activity information into AlphaEarth representations, with mobility serving as the dominant source of behavioral context.

\subsection{Downstream Prediction Performance}
Across the nine downstream tasks, BEACON exhibits a clear specialization toward human-centered prediction while remaining competitive on physical and environmental benchmarks (Tables~\ref{tab:reg}--\ref{tab:cls}). 

For the four human-related regression targets, BEACON achieves the best performance among all representations. Relative to AlphaEarth, $R^2$ improves from 0.379 to 0.463 for median household income (+22\%), from 0.431 to 0.614 for obesity prevalence (+43\%), and from 0.393 to 0.528 for poor mental health (+34\%). Nighttime lights also achieve the highest score (0.680 versus 0.663 for AlphaEarth). These gains are most pronounced for outcomes closely tied to collective human behavior and place usage.

In contrast, the ranking differs for physical and environmental targets. Clay achieves the strongest performance on daytime and nighttime land-surface temperature and PM$_{2.5}$, while TESSERA is generally second. BEACON performs nearly identically to AlphaEarth on these tasks. For example, PM$_{2.5}$ improves only marginally from 0.845 to 0.856 (+1.3\%), while LST performance remains essentially unchanged. This indicates that incorporating human context does not substantially alter the physical environmental information already encoded by Earth-observation foundation models.

The classification results reveal a contrasting pattern (Table~\ref{tab:cls}). For land-cover classification, which reflects physical surface morphology, AlphaEarth achieves the highest Macro-F1 (0.584), with BEACON second (0.567). In contrast, for functional land use, which is a human-defined construct, BEACON is best (Macro-F1 0.471 vs.\ 0.456), with the same ordering for accuracy. This land-cover/land-use contrast cleanly separates physical from human-defined structure. Similar patterns hold under nonlinear MLP probes, and the gains on human targets are most pronounced under the linear probe, indicating that the alignment primarily increases the linear decodability of human signal already latent in AlphaEarth. Ablations further confirm the complementary roles of semantic and behavioral supervision (Table~\ref{tab:ablation}). AE-Mobility and AE-POI both outperform AE-only on human-related prediction, while the full tri-modal BEACON model achieves the best average performance on both human-related and physical/environmental regression tasks.

\begin{table}[!htbp]
  \centering
  \small
  \caption{Classification performance (linear probe; mean\,$\pm$\,s.d.\ over five seeds). \textbf{Best} in bold, \underline{second-best} underlined.}
  \label{tab:cls}
  \setlength{\tabcolsep}{3.5pt}
  \begin{tabular}{@{}l cc cc@{}}
    \toprule
    & \multicolumn{2}{c}{Land cover} & \multicolumn{2}{c}{Functional land use} \\
    \cmidrule(lr){2-3}\cmidrule(lr){4-5}
    Model & Macro-F1 & Acc. & Macro-F1 & Acc. \\
    \midrule
    Coordinates & 0.152\,$\pm$\,0.001 & 0.266\,$\pm$\,0.007 & 0.144\,$\pm$\,0.007 & 0.241\,$\pm$\,0.007 \\
    Space2Vec   & 0.190\,$\pm$\,0.002 & 0.253\,$\pm$\,0.002 & 0.270\,$\pm$\,0.001 & 0.333\,$\pm$\,0.003 \\
    SatCLIP     & 0.230\,$\pm$\,0.002 & 0.338\,$\pm$\,0.005 & 0.153\,$\pm$\,0.002 & 0.252\,$\pm$\,0.008 \\
    TESSERA     & \underline{0.521\,$\pm$\,0.003} & \underline{0.635\,$\pm$\,0.004} & \underline{0.421\,$\pm$\,0.003} & 0.518\,$\pm$\,0.003 \\
    Clay        & 0.357\,$\pm$\,0.004 & 0.464\,$\pm$\,0.005 & 0.279\,$\pm$\,0.003 & 0.355\,$\pm$\,0.004 \\
    AlphaEarth     & \textbf{0.584\,$\pm$\,0.002} & \textbf{0.701\,$\pm$\,0.002} & 0.456\,$\pm$\,0.002 & \underline{0.549\,$\pm$\,0.002} \\
    \textbf{BEACON} & 0.567\,$\pm$\,0.002 & 0.687\,$\pm$\,0.002 & \textbf{0.471\,$\pm$\,0.002} & \textbf{0.565\,$\pm$\,0.002} \\
    \bottomrule
  \end{tabular}
\end{table}

\begin{table}[!htbp]
  \centering
  \small
  \caption{Ablation results of different modality combinations using linear probes.}
  \label{tab:ablation}
  \setlength{\tabcolsep}{6pt}
  \begin{tabular}{@{}lcc@{}}
    \toprule
    Variant &
    Human-related Avg. $R^2$ &
    Physical / Environmental Avg. $R^2$ \\
    \midrule
    Mobility only &
    0.011\,$\pm$\,0.030 &
    0.015\,$\pm$\,0.007 \\

    POI only &
    0.288\,$\pm$\,0.047 &
    0.199\,$\pm$\,0.009 \\

    Concat Raw &
    0.294\,$\pm$\,0.059 &
    0.644\,$\pm$\,0.007 \\

    AlphaEarth &
    0.401\,$\pm$\,0.037 &
    0.743\,$\pm$\,0.008 \\

    AE-Mobility &
    0.475\,$\pm$\,0.029 &
    0.736\,$\pm$\,0.006 \\

    AE-POI &
    \underline{0.494\,$\pm$\,0.023} &
    \underline{0.750\,$\pm$\,0.007} \\

    \textbf{BEACON} &
    \textbf{0.535\,$\pm$\,0.022} &
    \textbf{0.759\,$\pm$\,0.008} \\
    \bottomrule
  \end{tabular}
\end{table}

\section{Conclusion and Future Work}
We presented BEACON, a tri-modal contrastive framework that enriches AlphaEarth embeddings with semantic and behavioral urban context while retaining an image-only representation at deployment. Across nine downstream tasks, BEACON substantially improves human-centered prediction while largely preserving the physical and environmental information encoded in AlphaEarth.

More importantly, our findings suggest that physical appearance and urban function constitute distinct representational dimensions of geographic space. While current geospatial foundation models provide strong representations of the physical environment, aligning them with semantic and behavioral signals substantially improves their ability to represent human activity and urban function. Future work should evaluate the generality of this approach across cities, time periods, and mobility data sources, and explore incorporating behavioral alignment directly into foundation-model pretraining.

\bibliographystyle{ACM-Reference-Format}
\bibliography{cleaned_references}

@article{brown2025alphaearth,
  title   = {{AlphaEarth Foundations}: An Embedding Field Model for Accurate and Efficient Global Mapping from Sparse Label Data},
  author  = {Brown, Christopher F. and Kazmierski, Michal R. and Pasquarella, Valerie J. and Rucklidge, William J. and Samsikova, Masha and Zhang, Chenhui and Shelhamer, Evan and Lahera, Estefania and Wiles, Olivia and Ilyushchenko, Simon and Gorelick, Noel and Zhang, Lihui Lydia and Alj, Sophia and Schechter, Emily and Askay, Sean and Guinan, Oliver and Moore, Rebecca and Boukouvalas, Alexis and Kohli, Pushmeet},
  journal = {arXiv preprint arXiv:2507.22291},
  year    = {2025},
  doi     = {10.48550/arXiv.2507.22291},
  url     = {https://arxiv.org/abs/2507.22291}
}

@article{niu2021urbanfunctional,
  title     = {Delineating Urban Functional Use from Points of Interest Data with Neural Network Embedding: A Case Study in Greater London},
  author    = {Niu, Haifeng and Silva, Elisabete A.},
  journal   = {Computers, Environment and Urban Systems},
  volume    = {88},
  pages     = {101651},
  year      = {2021},
  publisher = {Elsevier},
  doi       = {10.1016/j.compenvurbsys.2021.101651}
}

@article{wang2025poi,
  title     = {Multi-Modal Contrastive Learning of Urban Space Representations from {POI} Data},
  author    = {Wang, Xinglei and Cheng, Tao and Law, Stephen and Zeng, Zichao and Yin, Lu and Liu, Junyuan},
  journal   = {Computers, Environment and Urban Systems},
  volume    = {120},
  pages     = {102299},
  year      = {2025},
  publisher = {Elsevier},
  doi       = {10.1016/j.compenvurbsys.2025.102299}
}

@article{liu2025beyondalphaearth,
  title   = {Beyond {AlphaEarth}: Toward Human-Centered Spatial Representation via {POI}-Guided Contrastive Learning},
  author  = {Liu, Junyuan and Qin, Quan and Dong, Guangsheng and Wang, Xinglei and Feng, Jiazhuang and Zeng, Zichao and Cheng, Tao},
  journal = {arXiv preprint arXiv:2510.09894},
  year    = {2025},
  doi     = {10.48550/arXiv.2510.09894},
  url     = {https://arxiv.org/abs/2510.09894}
}

@inproceedings{wen2026mora,
  title     = {{MoRA}: Mobility as the Backbone for Geospatial Representation Learning at Scale},
  author    = {Wen, Ya and Cai, Jixuan and Ma, Qiyao and Li, Linyan and Chen, Xinhua and Webster, Chris and Zhou, Yulun},
  booktitle = {International Conference on Learning Representations},
  year      = {2026},
  note      = {arXiv:2506.01297},
  doi       = {10.48550/arXiv.2506.01297},
  url       = {https://arxiv.org/abs/2506.01297}
}

\end{document}